\documentclass[11pt,a4paper]{article}
\usepackage[nohyperref]{acl2017}
\usepackage{times}
\usepackage{latexsym}

\usepackage{booktabs}
\usepackage{multirow}
\usepackage{graphicx}
\usepackage{color}
\usepackage{amsmath}
\usepackage{amssymb}

\usepackage{url}

\aclfinalcopy

\title{Sequential Attention:\\ A Context-Aware Alignment Function for Machine Reading}

\author{\bf{Sebastian Brarda\thanks{\hspace*{1ex} These authors contributed equally to this work.}}\\
Center for Data Science\\
New York University \\
\texttt{sb5518@nyu.edu}\\
\And
{\bf Philip Yeres\footnotemark[1]}  \\
Center for Data Science\\
New York University \\
\texttt{yeres@nyu.edu}\\ 
\And
{\bf Samuel R. Bowman}  \\
Center for Data Science\\
and Department of Linguistics\\
New York University \\
\texttt{bowman@nyu.edu}\\} 

\date{}

\begin{document}

\maketitle
 
\begin{abstract}
In this paper we  propose a neural network model with a novel \textit{Sequential Attention} layer that extends soft attention by assigning weights to words in an input sequence in a way that takes into account not just how well that word matches a query, but how well surrounding words match. We evaluate this approach on the task of reading comprehension (on the \textit{Who did What} and \textit{CNN} datasets) and show that it dramatically improves a strong baseline---the \textit{Stanford Reader}---and is competitive with the state of the art.
\end{abstract}

\section{Introduction}

Soft attention \cite{bahdanau2014neural}, a differentiable method for selecting the inputs for a component of a model from a set of possibilities, has been crucial to the success of artificial neural network models for natural language understanding tasks like reading comprehension that take short passages as inputs. However, standard approaches to attention in NLP select words with only very indirect consideration of their context, limiting their effectiveness. This paper presents a method to address this by adding explicit context sensitivity into the soft attention scoring function.

We demonstrate the effectiveness of this approach on the task of \textit{cloze}-style reading comprehension. A problem in the cloze style consists of a passage \textit{p}, a question \textit{q} and an answer \textit{a} drawn from among the entities mentioned in the passage. In particular, we use the  \textit{CNN} dataset \cite{hermann}, which introduced the task into widespread use in evaluating neural networks for language understanding, and the newer and more carefully quality-controlled \textit{Who did What} dataset \cite{onishi}.

\begin{figure}[t!]
\centering
\includegraphics[scale=1]{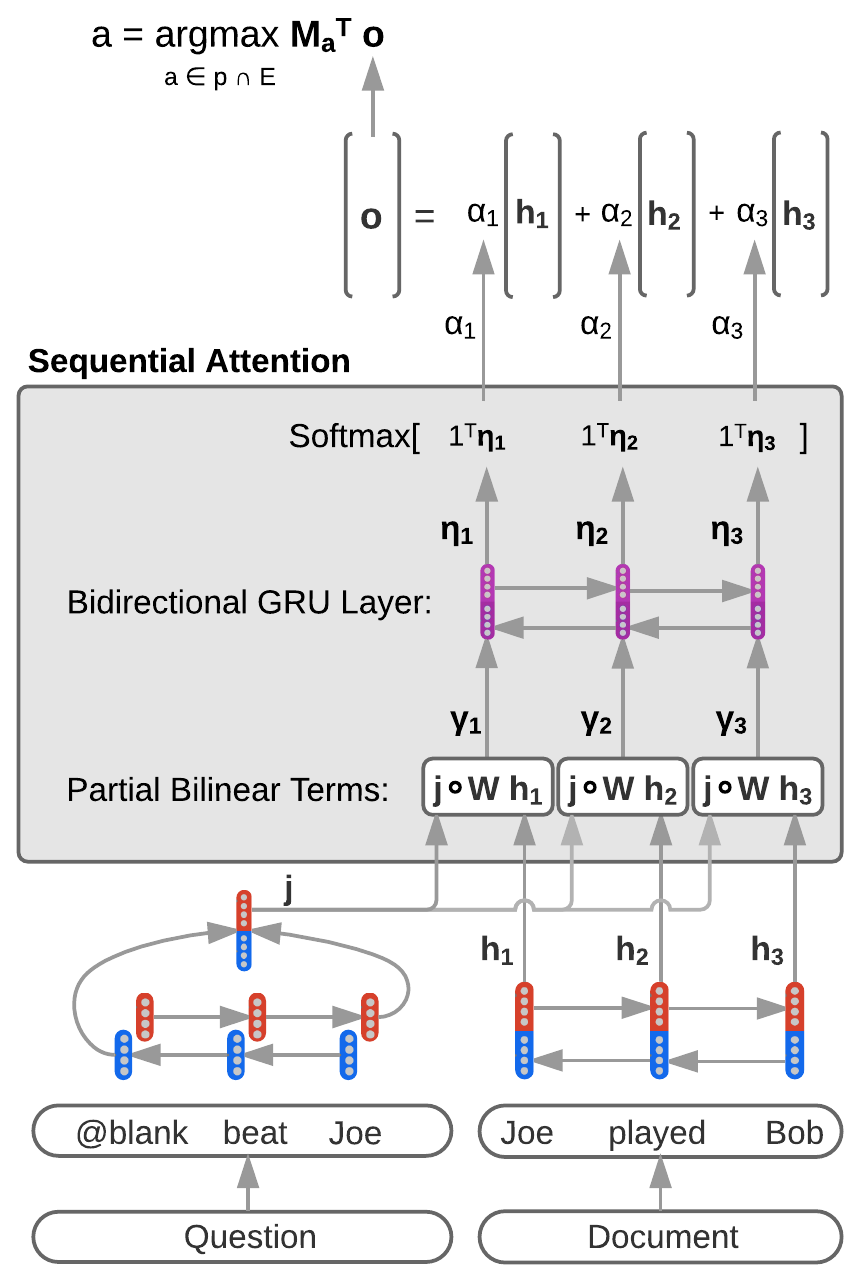}
\caption{The Sequential Attention Model. RNNs first encode the question into a vector $\boldsymbol{j}$ and the document into a sequence of vectors $\boldsymbol{H}$. For each word index $i$ in the document, a scoring vector $\boldsymbol{\gamma_i}$ is then computed from $\boldsymbol{j}$ and $\boldsymbol{h_i}$ using a function like the partial bilinear function shown here. These vectors are then used as inputs to another RNN layer, the outputs of which ($\boldsymbol{\eta_{i}}$) are summed elementwise and used as attention scores ($\alpha_{i}$) in answer selection.} 
\label{fig:mesh1}
\end{figure}

In standard approaches to soft attention over passages, a \textit{scoring function} is first applied to every word in the source text to evaluate how closely that word matches a  \textit{query vector} (here, a function of the question). The resulting scores are then normalized and used as the weights in a weighted sum which produces an \textit{output} or \textit{context} vector summarizing the most salient words of the input, which is then used in a downstream model (here, to select an answer).

In this work we propose a novel scoring function for soft attention that we call \textit{Sequential Attention} (SA), shown in Figure~\ref{fig:mesh1}. In an SA model, a mutiplicative interaction scoring function is used to produce a scoring \textit{vector} for each word in the source text. A newly-added bidirectional RNN then consumes those vectors and uses them to produce a context-aware scalar score for each word. We evaluate this scoring function within the context of the \textit{Stanford Reader} \cite{chen}, and show that it yields dramatic improvements in performance. On both datasets, it is outperformed only by the  \textit{Gated Attention Reader} \cite{DBLP:journals/corr/DhingraLCS16}, which in some cases has access to features not explicitly seen by our model.

\section{Related Work}
In addition to \newcite{chen}'s \textit{Stanford Reader} model, there have been several other modeling approaches developed to address these reading comprehension tasks. 

\newcite{DBLP:journals/corr/SeoKFH16} introduced the \textit{Bi-Directional Attention Flow} which consists of a multi-stage hierarchical process to represent context at different levels of granularity; it use the concatenation of passage word representation, question word representation, and the element-wise product of these vectors in their attention flow layer. This is a more complex variant of the classic bi-linear term that multiplies this concatenated vector with a vector of weights, producing attention scalars. 
\newcite{DBLP:journals/corr/DhingraLCS16}'s \textit{Gated-Attention Reader} integrates a multi-hop structure with a novel attention mechanism, essentially building query specific representations of the tokens in the document to improve prediction. This model conducts a classic dot-product soft attention to weight the query representations which are then multiplied element-wise with the context representations, and fed into the next layer of RNN. After several hidden layers that repeat the same process, the dot product between the context representation and the query is used to compute a classic soft-attention.

Outside the task of reading comprehension there has been other work on soft attention over text, largely focusing on the problem of attending over single sentences. \newcite{luong2015effective} study several issues in the design of soft attention models in the context of translation, and introduce the bilinear scoring function. They also propose the idea of attention input-feeding where the original attention vectors are concatenated with the hidden representations of the words and fed into the next RNN step. The goal is to make the model fully aware of the previous alignment choices. 

In work largely concurrent to our own, \newcite{kim2017structured} explore the use of conditional random fields (CRFs) to impose a variety of constraints on attention distributions achieving strong results on several sentence level tasks.

\section{Modeling}

Given the tuple (passage, question, answer), our goal is to predict $\Pr(a|d,q)$ where $a$ refers to answer, $d$ to passage, and $q$ to question. We define the words of each passage and question as $d={d_1,..,d_m}$ and $q={q_1,...,q_l}$, respectively, where exactly one $q_i$ contains the token \textit{@blank}, representing a blank that can be correctly filled in by the answer. With calibrated probabilities $\Pr(a|d,q)$, we take the ${\operatornamewithlimits{argmax}_a} \Pr(a|d,q)$ where possible $a$'s are restricted to the subset of anonymized entity symbols present in $d$.
In this section, we present two models for this reading comprehension task: \newcite{chen}'s \textit{Stanford Reader}, and our version with a novel attention mechanism which we call the \textit{Sequential Attention} model.

\subsection{Stanford Reader}
\paragraph{Encoding}
Each word or entity symbol is mapped to a d-dimensional vector via embedding matrix $E \in \mathbb{R}^{d \times |V|}$. For simplicity, we denote the vectors of the passage and question as $\boldsymbol{d={d_1,..,d_m}}$ and $\boldsymbol{q={q_1,...,q_l}}$, respectively. The \textit{Stanford Reader} \cite{chen} uses bidirectional GRUs \cite{cho2014learning} to encode the passage and questions. For the passage, the hidden state is defined: $
\boldsymbol{h_i}=\textrm{concat}(\overrightarrow{\boldsymbol{h_i}},\overleftarrow{\boldsymbol{h_i}} )$.
Where contextual embeddings $\boldsymbol{d_i}$ of each word in the passage are encoded in both directions.
\begin{equation}
\overleftarrow{\boldsymbol{h_i}} = \textrm{GRU}(\boldsymbol{\overleftarrow{h_{i+1}}}, \boldsymbol{d_i)}
\end{equation}
\vspace{-1.0em}
\begin{equation}
\overrightarrow{\boldsymbol{h_i}} = \textrm{GRU}(\boldsymbol{\overrightarrow{h_{i-1}}},\boldsymbol{d_i})
\end{equation}
And for the question, the last hidden representation of each direction is concatenated: \begin{equation}
\boldsymbol{j}=\textrm{concat}(\boldsymbol{\overrightarrow{j_l}},\boldsymbol{\overleftarrow{j_1}})
\end{equation}

\paragraph{Attention and answer selection}
The Stanford Reader uses bilinear attention \cite{luong2015effective}:
\begin{equation}
\alpha_i=\textrm{softmax}_i(\boldsymbol{j}\mathbf{W}\boldsymbol{h_i})
\end{equation}
Where $\mathbf{W}$ is a learned parameters matrix of the bilinear term that computes the similarity between $\boldsymbol{j}$ and $\boldsymbol{h_i}$ with greater flexibility than a dot product. The output vector is then computed as a linear combination of the hidden representations of the passage, weighted by the attention coefficients:
\begin{equation}
\boldsymbol{o}=\sum \alpha_i \boldsymbol{h_i}
\end{equation}
The prediction is the answer, $a$, with highest probability from among the anonymized entities:
\begin{equation}
a={\operatornamewithlimits{argmax}_{a \in p \cap entities}} M_a^T\boldsymbol{o}
\end{equation}
Here, $\mathbf{M}$ is the weight matrix that maps the output to the entities, and $M_a$ represents the column of a certain entity. Finally a softmax layer is added on top of $M_a^T\boldsymbol{o}$ with a negative log-likelihood objective for training.

\subsection{Sequential Attention}
In the \textit{Sequential Attention} model instead of producing a single scalar value $\alpha_i$ for each word in the passage by using a bilinear term, we define the vectors  $\boldsymbol{\gamma_i}$ with a partial-bilinear term\footnote{Note that doing softmax over the sum of the terms of the $\gamma_i$ vectors would lead to the same $\alpha_i$ of the Stanford Reader.}. Instead of doing the dot product as in the bilinear term, we conduct an element wise multiplication to produce a vector instead of a scalar:
\begin{equation}
\boldsymbol{\gamma_i} = \boldsymbol{j} \circ \mathbf{W} \boldsymbol{h_i}
\end{equation}
Where $\mathbf{W}$ is a matrix of learned parameters. It is also possible to use an element-wise multiplication, thus prescinding
the parameters $\mathbf{W}$:
\begin{equation}
\boldsymbol{\gamma_i} = \boldsymbol{j} \circ \boldsymbol{h_i}
\end{equation}
 We then feed the $\boldsymbol{\gamma_i}$ vectors into a new bidirectional GRU layer to get the hidden attention $\boldsymbol{\eta_i}$ vector representation.
 \begin{equation}
 \boldsymbol{\overleftarrow{\eta_i}} = \textrm{GRU}(\boldsymbol{\overleftarrow{\eta_{i+1}}}, \boldsymbol{\gamma_i})
 \end{equation}
 \vspace{-1.0em}
 \begin{equation}
 \boldsymbol{\overrightarrow{\eta_i}} = \textrm{GRU}(\boldsymbol{\overrightarrow{\eta_{i-1}}}, \boldsymbol{\gamma_i})
 \end{equation}
We concatenate the directional $\boldsymbol{\eta}$ vectors to be consistent with the structure of previous layers.
\begin{equation}
\boldsymbol{\eta_i}=\textrm{concat}(\boldsymbol{\overrightarrow{\eta_i}}, \boldsymbol{\overleftarrow{\eta_i}})
\end{equation}
Finally, we compute the $\boldsymbol{\alpha}$ weights as below, and proceed as before.
\begin{equation}
\alpha_i=\textrm{softmax}_i(1^\top \boldsymbol{\eta_i}])
\end{equation}

\begin{equation}
\boldsymbol{o}=\sum \alpha_i \boldsymbol{h_i}
\end{equation}

\begin{equation}
a={\operatornamewithlimits{argmax}_{a \in p \cap entities}} M_a^T\boldsymbol{o}
\end{equation}

\section{Experiments and Results}

We evaluate our model on two tasks, \textit{CNN} and Who did What (\textit{WDW}). For CNN, we used the anonymized version of the dataset released by \newcite{hermann}, containing $380{,}298$ training, $3{,}924$ dev, and $3{,}198$ test examples. For \textit{WDW} we used \newcite{onishi}'s data generation script to reproduce their \textit{WDW} data, yielding $127{,}786$ training, $10{,}000$ dev, and $10{,}000$ test examples.\footnote{In the WDW data we found 340 examples in the strict training set, 545 examples in the relaxed training set, 20 examples in the test set, and 30 examples in the validation set that were not answerable because the anonymized answer entity did not exist in the passage. We removed these examples, reducing the size of the WDW test set by $0.2\%$, to $9{,}980$. We believe this difference is not significant and did not bias the comparison between models.} We used the strict version of \textit{WDW}.

\paragraph{Training}
We implemented all our models in Theano \cite{2016arXiv160502688short} and Lasagne \cite{lasagne} and used the Stanford Reader \cite{chen} open source implementation as a reference.
We largely used the same hyperparameters as \newcite{chen} in the Stanford Reader: $|V|=50K$, embedding size $d=100$,  \textit{GloVe} \cite{pennington2014glove} word embeddings\footnote{The GloVe word vectors used were pretrained with 6 billion tokens with an uncased vocab of 400K words, and were obtained from Wikipedia 2014 and Gigaword 5.} for initialization, hidden size $h=128$. The size of the hidden layer of the bidirectional RNN used to encode the attention vectors is double the size of the one that encodes the words, since it receives vectors that result from the concatenation of GRUs that go in both directions, $\boldsymbol{\eta} \in \mathbb{R}^{256}$.  Attention and output parameters were initialized from a $U \sim (-0.01,0.01)$ while GRU weights were initialized from a $N \sim (0,0.1)$. Learning was carried out with SGD with a learning rate of 0.1, batch size of 32, gradient clipping of norm 10 and dropout of $0.2$ in all the vertical layers\footnote{We also tried increasing the hidden size to 200, using 200d GloVe word representations and increasing the dropout rate to 0.3. Finally we increased the number of hidden encoding layers to two. None of these changes resulted in significant performance improvements in accordance with \newcite{chen}.} (including the \textit{Sequential Attention} layer). Also, all the anonymized entities were relabeled according to the order of occurrence, as in the \textit{Stanford Reader}. We trained all models for 30 epochs.

{\footnotesize
\begin{table}[t]
\centering
\label{results}
\begin{tabular}{@{}lrr@{}}
\toprule
\textbf{Model}                & \textbf{\textit{WDW} Strict}  & \textbf{\textit{CNN}}  \\ \midrule
Attentive Reader         &  53\%                 &      63\% \\
Stanford Reader  & 65.6\%     & 73.4\% \\
~~+ SA partial-bilinear & 67.2\%     & 77.1\% \\ 
Gated Att. Reader & \textbf{71.2}\%         & \textbf{77.9}\% \\ 
\bottomrule
\end{tabular}
\caption{Accuracy on \textit{WDW} and CNN test sets}
\vspace{-0.5em}
\end{table}
}

\begin{figure*}[t]
\graphicspath{ {img/} }
\centering
\includegraphics[scale=0.2975]{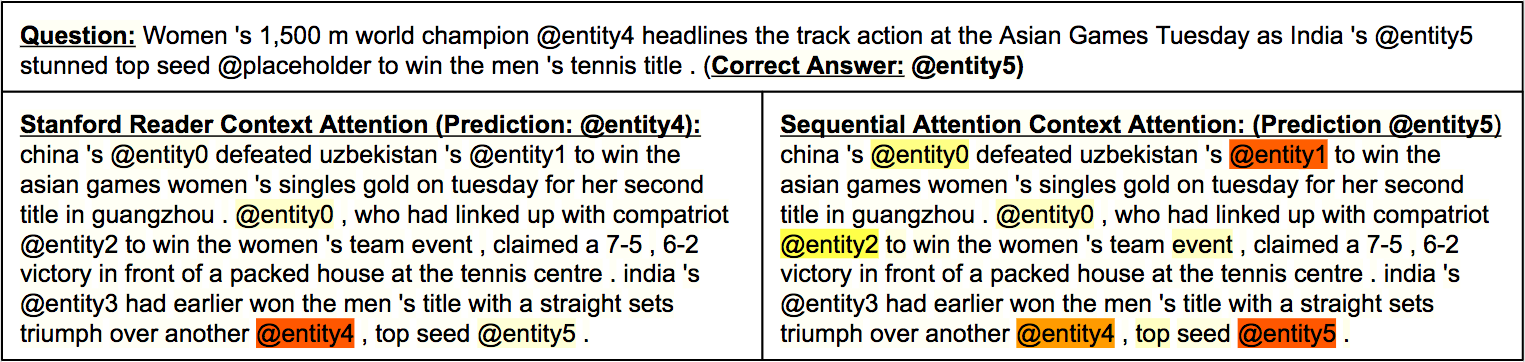}
\caption{Representative sample output for the Stanford Reader and our model. \label{fig:example}}
\end{figure*}

\subsection{Results}

\paragraph{Who did What}
In our experiments the \textit{Stanford Reader} (SR) achieved an accuracy of $65.6\%$ on the strict \textit{WDW} dataset compared to the $64\%$ that \newcite{onishi} reported. The \textit{Sequential Attention} model (SA) with partial-bilinear scoring function got  $67.21\%$, which is the second best performance on the $WDW$ leaderboard, only surpassed by the $71.2\%$ from the \textit{Gated Attention Reader} (GA) with \textit{qe-comm} \cite{DBLP:journals/corr/LiLHWCZX16} features and fixed GloVe embeddings. However, the GA model without \textit{qe-comm} features and fixed embeddings performs significantly worse at $67\%$. We did not use these features in our SA models, and it is likely that adding these features could further improve SA model performance. We also experimented with fixed embeddings in SA models, but fixed embeddings reduced SA performance.

Another experiment we conducted was to add $100$K training samples from \textit{CNN} to the \textit{WDW} data. This increase in the training data size boosted accuracy by $1.4\%$ with the SR and $1.8\%$ with the \textit{Sequential Attention} model reaching a $69\%$ accuracy. This improvement strongly suggests that the gap in performance/difficulty between the CNN and the \textit{WDW} datasets is partially related to the difference in the training set sizes which results in overfitting.

\paragraph{CNN}
For a final sanity check and a fair comparison against a well known benchmark, we ran our \textit{Sequential Attention} model on exactly the same \textit{CNN} data used by \newcite{chen}.

The \textit{Sequential Attention} model with partial-bilinear attention scoring function took an average of 2X more time per epoch to train vs. the \textit{Stanford Reader}. However, our model converged in only 17 epochs vs. 30 for the \textit{SR}. The results of training the \textit{SR} on \textit{CNN} were slightly lower than the $73.6\%$ reported by \newcite{chen}. The \textit{Sequential Attention} model achieved $77.1\%$ accuracy, a $3.7\%$ gain with respect to \textit{SR}.

\subsubsection{Model comparison on CNN}
After achieving good performance with \textit{SA} we wanted to understand what was driving the increase in accuracy. It is clear that \textit{SA} has more trainable parameters compared to \textit{SR}. However, it was not clear if the additional computation required to learn those parameters should be allocated in the attention mechanism, or used to compute richer hidden representations of the passage and questions. Additionally, the bilinear parameters increase the computational requirements, but their impact on performance was not clear. To answer these questions we compared the following models: i) \textit{SR} with dot-product attention; ii) \textit{SR} with bilinear attention; iii) \textit{SR} with two layers (to compute the hidden question and passage representations) and dot-product attention; iv) \textit{SR} with two layers and bilinear attention; v) \textit{SA} with elementwise multiplication scoring function; vi) \textit{SA} with partial-bilinear scoring function.

Surprisingly, the element-wise version of \textit{SA} performed better than the partial-bilinear version, with an accuracy of $77.3\%$ which, to our knowledge, has only been surpassed by \newcite{DBLP:journals/corr/DhingraLCS16} with their \textit{Gated-Attention Reader} model. 

Additionally, 1-layer \textit{SR} with dot-product attention got $0.3\%$ lower accuracy than the 1-layer \textit{SR} with bilinear attention. These results suggest that the bilinear parameters do not significantly improve performance over dot-product attention.

Adding an additional GRU layer to encode the passage and question in the \textit{SR} model increased performance over the original 1-layer model. With dot-product attention the increase was $1.1\%$ whereas with bilinear attention, the increase was $1.3\%$. However, these performance increases were considerably less than the lift from using an \textit{SA} model (and SA has fewer parameters).

{\footnotesize
\begin{table}[t]
\centering
\label{results2}
\begin{tabular}{@{}lrr@{}}
\toprule
\textbf{Model} &  \textbf{\textit{CNN}} & \textbf{Params}\\ \midrule
SR, dot prod. att.             &  73.1\% &      $5.44 \times 10^{6}$\\
SR, bilinear att.              &  73.4\%  &      $5.50 \times 10^{6}$\\
SR, 2-layer, dot prod. att.    &  74.2\%  &      $5.83 \times 10^{6}$\\
SR, 2-layer, bilinear att.     &  74.7\%  &      $5.90 \times 10^{6}$\\
SA, element-wise att.          &  77.3\%  &      $5.73 \times 10^{6}$\\ 
SA, partial-bilinear att.      &  77.1\%  &      $5.80 \times 10^{6}$\\ 
\bottomrule
\end{tabular}
\caption{Accuracy on CNN test sets and number of trainable parameters for various Stanford Reader (SR) and Sequential Attention (SA) models.}
\end{table}
}

\subsection{Discussion}

The difference between our \textit{Sequential Attention} and standard approaches to attention is that we conserve the distributed representation of similarity for each token and use that contextual information when computing attention over other words. In other words, when the bilinear attention layer computes $\alpha_i=\textrm{softmax}_i(\boldsymbol{j}\mathbf{W}\boldsymbol{h_i})$, it only cares about the magnitude of the resulting $\alpha_i$ (the amount of attention that it gives to that word). Whereas if we keep the vector $\boldsymbol{\gamma_i}$ we can also know which were the dimensions of the distributed representation of the attention that weighted in that decision. Furthermore, if we use that information to feed a new GRU, it helps the model to learn how to assign attention to surrounding words.

Compared to \textit{Sequential Attention}, \textit{Bidirectional attention flow} uses a considerably more complex architecture with a query representations for each word in the question. Unlike the \textit{Gated Attention Reader}, SA does not require intermediate soft attention and it uses only one additional RNN layer. Furthermore, in SA no dot product is required to compute attention, only the sum of the elements of the $\boldsymbol{\eta}$ vector. SA's simpler architecture performs close to the state-of-the-art.

Figure \ref{fig:example} shows some sample model behavior. In this example and elsewhere, \textit{SA} results in less sparse attention vectors compared to \textit{SR}, and this helps the model assign attention not only to potential target strings (anonymized entities) but also to relevant contextual words that are related to those entities. This ultimately leads to richer semantic representations $\boldsymbol{o}=\sum \alpha_i \boldsymbol{h_i}$ of the passage. 

Finally, we found: i) bilinear attention does not yield dramatically higher performance compared to dot-product attention; ii) bilinear parameters do not improve SA performance; iii) Increasing the number of layers in the attention mechanism yields considerably greater performance gains with fewer parameters compared to increasing the number of layers used to compute the hidden representations of the question and passage.

\section{Conclusion and Discussion}
In this this paper we created a novel and simple model with a \textit{Sequential Attention} mechanism that performs near the state of the art on the \textit{CNN} and \textit{WDW} datasets by improving the bilinear and dot-product attention mechanisms with an additional bi-directional RNN layer. This additional layer allows local alignment information to be used when computing the attentional score for each token. Furthermore, it provides higher performance gains with fewer parameters compared to adding an additional layer to compute the question and passage hidden representations.
For future work we would like to try other machine reading datasets such as \textit{SQuAD} and \textit{MS MARCO}. Also, we think that some elements of the \textit{SA} model could be mixed with ideas applied in recent research from \newcite{DBLP:journals/corr/DhingraLCS16} and \newcite{DBLP:journals/corr/SeoKFH16}. We believe that the \textit{SA} mechanism may benefit other tasks as well, such as machine translation.

\section*{Acknowledgements}
This paper was the result of a term project for the NYU Course DS-GA 3001, Natural Language Understanding with Distributed Representations. Bowman acknowledges support from a Google Faculty Research Award and gifts from Tencent Holdings and the NVIDIA Corporation.

\bibliography{acl2017}

\begin{thebibliography}{}
\expandafter\ifx\csname natexlab\endcsname\relax\def\natexlab#1{#1}\fi

\bibitem[{Bahdanau et~al.(2014)Bahdanau, Cho, and Bengio}]{bahdanau2014neural}
Dzmitry Bahdanau, Kyunghyun Cho, and Yoshua Bengio. 2014.
\newblock \href{http://arxiv.org/abs/1409.0473}{Neural machine translation by
  jointly learning to align and translate}.
\newblock {\em CoRR\/} abs/1409.0473.
\newblock
  \href{http://arxiv.org/abs/1409.0473}{http://arxiv.org/abs/1409.0473}.

\bibitem[{Chen et~al.(2016)Chen, Bolton, and Manning}]{chen}
Danqi Chen, Jason Bolton, and Christopher~D. Manning. 2016.
\newblock \href{http://arxiv.org/abs/1606.02858}{A thorough examination of the
  cnn/daily mail reading comprehension task}.
\newblock {\em CoRR\/} abs/1606.02858.
\newblock
  \href{http://arxiv.org/abs/1606.02858}{http://arxiv.org/abs/1606.02858}.

\bibitem[{Cho et~al.(2014)Cho, van Merrienboer, G{\"{u}}l{\c{c}}ehre, Bougares,
  Schwenk, and Bengio}]{cho2014learning}
Kyunghyun Cho, Bart van Merrienboer, {\c{C}}aglar G{\"{u}}l{\c{c}}ehre, Fethi
  Bougares, Holger Schwenk, and Yoshua Bengio. 2014.
\newblock \href{http://arxiv.org/abs/1406.1078}{Learning phrase representations
  using {RNN} encoder-decoder for statistical machine translation}.
\newblock {\em CoRR\/} abs/1406.1078.
\newblock
  \href{http://arxiv.org/abs/1406.1078}{http://arxiv.org/abs/1406.1078}.

\bibitem[{Dhingra et~al.(2016)Dhingra, Liu, Cohen, and
  Salakhutdinov}]{DBLP:journals/corr/DhingraLCS16}
Bhuwan Dhingra, Hanxiao Liu, William~W. Cohen, and Ruslan Salakhutdinov. 2016.
\newblock \href{http://arxiv.org/abs/1606.01549}{Gated-attention readers for
  text comprehension}.
\newblock {\em CoRR\/} abs/1606.01549.
\newblock
  \href{http://arxiv.org/abs/1606.01549}{http://arxiv.org/abs/1606.01549}.

\bibitem[{Dieleman et~al.(2015)Dieleman, Schlüter, Raffel, Olson, Sønderby,
  Nouri, Maturana, Thoma, Battenberg, Kelly, Fauw, Heilman, de~Almeida, McFee,
  Weideman, Takács, de~Rivaz, Crall, Sanders, Rasul, Liu, French, and
  Degrave}]{lasagne}
Sander Dieleman, Jan Schlüter, Colin Raffel, Eben Olson, Søren~Kaae
  Sønderby, Daniel Nouri, Daniel Maturana, Martin Thoma, Eric Battenberg, Jack
  Kelly, Jeffrey~De Fauw, Michael Heilman, Diogo~Moitinho de~Almeida, Brian
  McFee, Hendrik Weideman, Gábor Takács, Peter de~Rivaz, Jon Crall, Gregory
  Sanders, Kashif Rasul, Cong Liu, Geoffrey French, and Jonas Degrave. 2015.
\newblock \href{https://doi.org/10.5281/zenodo.27878}{Lasagne: First release.}
\newblock
  \href{https://doi.org/10.5281/zenodo.27878}{https://doi.org/10.5281/zenodo.27878}.

\bibitem[{Hermann et~al.(2015)Hermann, Kocisk{\'{y}}, Grefenstette, Espeholt,
  Kay, Suleyman, and Blunsom}]{hermann}
Karl~Moritz Hermann, Tom{\'{a}}s Kocisk{\'{y}}, Edward Grefenstette, Lasse
  Espeholt, Will Kay, Mustafa Suleyman, and Phil Blunsom. 2015.
\newblock \href{http://arxiv.org/abs/1506.03340}{Teaching machines to read and
  comprehend}.
\newblock {\em CoRR\/} abs/1506.03340.
\newblock
  \href{http://arxiv.org/abs/1506.03340}{http://arxiv.org/abs/1506.03340}.

\bibitem[{Kim et~al.(2017)Kim, Denton, Hoang, and Rush}]{kim2017structured}
Yoon Kim, Carl Denton, Luong Hoang, and Alexander~M. Rush. 2017.
\newblock \href{http://arxiv.org/abs/1702.00887}{Structured attention
  networks}.
\newblock {\em CoRR\/} abs/1702.00887.
\newblock
  \href{http://arxiv.org/abs/1702.00887}{http://arxiv.org/abs/1702.00887}.

\bibitem[{Li et~al.(2016)Li, Li, He, Wang, Cao, Zhou, and
  Xu}]{DBLP:journals/corr/LiLHWCZX16}
Peng Li, Wei Li, Zhengyan He, Xuguang Wang, Ying Cao, Jie Zhou, and Wei Xu.
  2016.
\newblock \href{http://arxiv.org/abs/1607.06275}{Dataset and neural recurrent
  sequence labeling model for open-domain factoid question answering}.
\newblock {\em CoRR\/} abs/1607.06275.
\newblock
  \href{http://arxiv.org/abs/1607.06275}{http://arxiv.org/abs/1607.06275}.

\bibitem[{Luong et~al.(2015)Luong, Pham, and Manning}]{luong2015effective}
Minh{-}Thang Luong, Hieu Pham, and Christopher~D. Manning. 2015.
\newblock \href{http://arxiv.org/abs/1508.04025}{Effective approaches to
  attention-based neural machine translation}.
\newblock {\em CoRR\/} abs/1508.04025.
\newblock
  \href{http://arxiv.org/abs/1508.04025}{http://arxiv.org/abs/1508.04025}.

\bibitem[{Onishi et~al.(2016)Onishi, Wang, Bansal, Gimpel, and
  McAllester}]{onishi}
Takeshi Onishi, Hai Wang, Mohit Bansal, Kevin Gimpel, and David~A. McAllester.
  2016.
\newblock \href{http://arxiv.org/abs/1608.05457}{Who did what: {A} large-scale
  person-centered cloze dataset}.
\newblock {\em CoRR\/} abs/1608.05457.
\newblock
  \href{http://arxiv.org/abs/1608.05457}{http://arxiv.org/abs/1608.05457}.

\bibitem[{Pennington et~al.(2014)Pennington, Socher, and
  Manning}]{pennington2014glove}
Jeffrey Pennington, Richard Socher, and Christopher~D. Manning. 2014.
\newblock \href{http://www.aclweb.org/anthology/D14-1162}{Glove: Global vectors
  for word representation}.
\newblock In {\em Empirical Methods in Natural Language Processing (EMNLP)\/}.
  pages 1532--1543.
\newblock
  \href{http://www.aclweb.org/anthology/D14-1162}{http://www.aclweb.org/anthology/D14-1162}.

\bibitem[{Seo et~al.(2016)Seo, Kembhavi, Farhadi, and
  Hajishirzi}]{DBLP:journals/corr/SeoKFH16}
Min~Joon Seo, Aniruddha Kembhavi, Ali Farhadi, and Hannaneh Hajishirzi. 2016.
\newblock \href{http://arxiv.org/abs/1611.01603}{Bidirectional attention flow
  for machine comprehension}.
\newblock {\em CoRR\/} abs/1611.01603.
\newblock
  \href{http://arxiv.org/abs/1611.01603}{http://arxiv.org/abs/1611.01603}.

\bibitem[{{Theano Development Team}(2016)}]{2016arXiv160502688short}
{Theano Development Team}. 2016.
\newblock \href{http://arxiv.org/abs/1605.02688}{Theano: {A} python framework
  for fast computation of mathematical expressions}.
\newblock {\em CoRR\/} abs/1605.02688.
\newblock
  \href{http://arxiv.org/abs/1605.02688}{http://arxiv.org/abs/1605.02688}.

\end{thebibliography}
\bibliographystyle{acl_natbib}

\end{document}